\documentclass[journal]{IEEEtran}
\usepackage{amsmath,amsfonts}
\usepackage{algorithmic}
\usepackage{algorithm}
\usepackage{array}
\usepackage{textcomp}
\usepackage{stfloats}
\usepackage{url}
\usepackage{graphicx}
\usepackage{cite}
\usepackage{orcidlink}

\begin{document}

\title{Maze Runner: Autonomous Maze Navigation with Line-Based Localization and Control Barrier Functions}

\author{Alvaro J. Gaona \orcidlink{0009-0003-4967-4444}, Lucas Gómez-Velayos \orcidlink{0009-0005-3829-8361}, Peter Pasuy-Quevedo \orcidlink{0009-0003-8768-0685}
\thanks{Code: \url{https://github.com/alvgaona/maze-runner}}}

\maketitle

\begin{abstract}
Maze navigation requires localization from sparse, repetitive geometry, planning through narrow corridors, and control that stays safe when the estimate or the map is wrong. We present Maze Runner, an open autonomy stack in C++ and ROS~2 that addresses these needs end to end on a simulated Husarion ROSbot XL with a 2D LiDAR in Gazebo. Perception extracts wall lines from each scan with RANSAC and total least squares, with per-line uncertainty. The same line front end feeds two estimators: an Extended Kalman Filter (EKF) that localizes against a line map built from the maze mesh, and an EKF SLAM variant that builds the line map online. Planning runs online on an inflated occupancy grid with A* plus line-of-sight shortcutting, or with a forward-only Hybrid A*. A nonlinear Model Predictive Controller (NMPC) tracks the plan while discrete-time Control Barrier Functions (CBFs) on world-frame LiDAR points keep the robot clear of walls and unmapped obstacles. The algorithms live in a ROS-free, unit-tested core library, and the whole stack builds and runs from a single reproducible environment. In simulation, the robot drives a 135~m route through the maze with 1.0~cm localization RMSE against the known map and 24~cm in SLAM mode, the planner returns a path in 6~ms, and the NMPC solves in 11.7~ms on average, never exceeding its 80~ms budget.
\end{abstract}


\section{Introduction}
\IEEEPARstart{A}{utonomous} mobile robots have attracted sustained interest over the past three decades as they tackle tasks previously beyond practical reach. Representative applications include inspection of electrical grids, thermal anomaly detection on photovoltaic components, package delivery, and exploration of unstructured environments. These and many related use cases continue to drive research and industrial development in this subfield of robotics. \\
\indent Unmanned Aerial Vehicles (UAVs), commonly known as drones, have become a central focus of robotics research, enabling numerous applications thanks to their agile dynamics and maneuverability. In particular, nano-drones operating as swarms have shown promise for exploration in unstructured environments \cite{10734219}. The energy industry is also increasingly automating key sectors, including solar power generation, where photovoltaic (PV) modules require regular maintenance and robotic platforms can survey large arrays of modules at lower cost \cite{6828739}. Drones have likewise emerged as a promising solution for power line inspection, reducing worker exposure to hazardous conditions while improving operational efficiency \cite{10341871}. \\
\indent Autonomous mobile robots (AMRs) operating in constrained environments such as mazes, corridors, and indoor spaces must address three interconnected challenges: perceiving sparse and often ambiguous geometry, planning feasible motions through narrow passages, and executing control robustly despite process and measurement uncertainty. Localization must run in real time without exception; continuous pose estimation is what lets the robot follow a plan safely. Planning, in turn, must be fast enough to replan whenever the goal changes. \\
\indent In this article, we present a maze navigation stack built on ROS~2 \cite{macenski2022ros2} and Gazebo \cite{koenig2004gazebo} that runs in real time and needs no infrastructure in the environment. The stack integrates: (i) geometric perception via RANSAC line extraction with total least squares refinement; (ii) line-based EKF localization against a map derived from the maze mesh, and an EKF SLAM mode that builds that map online; (iii) online grid planning with A* and line-of-sight shortcutting, or Hybrid A*; and (iv) path tracking with NMPC constrained by discrete-time CBFs \cite{9483029} on LiDAR points. An earlier version of this work was prototyped in MATLAB; this paper describes the redesigned C++ implementation and evaluates it in Gazebo.

Concisely, our contributions are outlined as follows:

\begin{enumerate}
    \item An open ROS~2 implementation whose algorithms live in a ROS-free, unit-tested C++ library, built and run from a single reproducible environment.
    \item A shared line front end (RANSAC with total least squares and per-line uncertainty) that serves both EKF localization against a mesh-derived line map and EKF SLAM with line landmarks.
    \item An NMPC with discrete-time CBF constraints on world-frame LiDAR points, with a sector-based obstacle selection that bounds the problem size, solved in real time.
    \item A comparison of online grid planners (A*, A* with shortcutting, and Hybrid A*) on the maze.
\end{enumerate}

The remainder of the paper reviews related work, describes the planning, mapping, localization, SLAM, and control layers, details the simulated platform, and reports experiments in Gazebo.

\section{Related Work}

\subsection{State Estimation and SLAM}

State estimation is fundamental to robot localization. The Extended Kalman Filter (EKF) remains one of the most widely adopted approaches, along with numerous variants derived from its original formulation. Teslić et al. \cite{Tesli2010} demonstrated EKF localization of a wheeled robot in structured environments using line features, modeling the system dynamics in the prediction step and incorporating sensor observations for state correction; the authors noted that extending this framework to Simultaneous Localization and Mapping (SLAM) would be a natural progression. EKF SLAM with geometric features has a long history: Castellanos et al. \cite{castellanos1999spmap} proposed the SPmap, a probabilistic framework for map building with segment features, and Dissanayake et al. \cite{dissanayake2001slam} established the convergence properties of the EKF SLAM solution. Cuadra-Troncoso et al. \cite{CUADRATRONCOSO2015219} proposed Polar Scan-Matching with Kalman Segmentation (PSM-KS), which estimates line segments in polar coordinates and builds a map from them.

\subsection{Line Fitting and Extraction}

Pfister et al.~\cite{1241772} fitted lines to two-dimensional laser scans with weighted least squares and used $\chi^2$ tests to match observed lines with the map while building it incrementally. Nguyen et al.~\cite{1545234} compared split-and-merge, incremental, Hough transform, line regression, RANSAC \cite{fischler1981ransac}, and Expectation-Maximization (EM) methods, and identified split-and-merge as the best-performing algorithm for real-time applications. We use RANSAC for its simplicity and robustness to clutter, followed by total least squares on the inliers.

\subsection{Path Planning}

A* \cite{hart1968astar} remains the standard for shortest paths on grids. Its 8-connected paths follow the grid directions, so post-processing such as line-of-sight shortcutting is commonly applied to obtain shorter, straighter paths. Hybrid A* \cite{dolgov08gppSTAIR, 6309512} searches over continuous poses with motion primitives that respect a minimum turning radius, which yields kinematically feasible paths for car-like robots in constrained environments.

\subsection{Control Barrier Functions}

CBFs add a layer of safety to optimization-based controllers, a property that is difficult to encode with conventional control techniques. Zeng et al. \cite{9483029} formulated discrete-time CBFs as constraints inside an MPC, so that safety is enforced along the predicted trajectory rather than only at the current state, and demonstrated it on obstacle avoidance and autonomous racing. Xiao and Belta \cite{9516971} introduced high-order CBFs (HOCBFs) for constraints whose relative degree is greater than one. Zheng et al.\ \cite{zheng2025nmpcblightweightsafetycriticalmotion} combined a lightweight neural-network planner with an MPC formulation and a CBF to ensure safety while following a path, using a slack variable in the CBF constraint and a time-varying scalar multiplier computed at each step.

\section{Methodology}

\begin{figure}[!t]
\centering
\includegraphics[width=\linewidth]{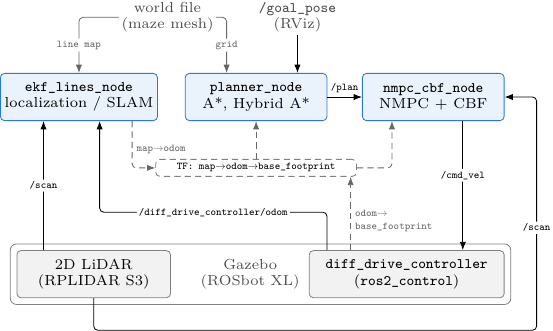}
\caption{Architecture of the navigation stack. Each layer is a ROS~2 node that wraps a ROS-free core library. Solid arrows are topics and dashed arrows TF transforms (the EKF publishes \texttt{map}$\to$\texttt{odom}, the drive controller \texttt{odom}$\to$\texttt{base\_footprint}); the line map and the occupancy grid are both built from the world file at startup.}
\label{fig:arch}
\end{figure}

This work addresses autonomous point-to-point navigation through a maze of planar walls: the robot must reach a goal from its current pose, and a new goal may be issued at any time. Fig.~\ref{fig:arch} shows the pipeline. A planning layer computes a collision-free path on an occupancy grid of the maze. A localization layer estimates the robot's pose by matching lines extracted from the LiDAR scan with a line map of the walls, or builds that map online in SLAM mode. A control layer tracks the plan with an NMPC whose CBF constraints keep the robot away from every LiDAR return, including obstacles that are not in the map. Both the occupancy grid and the line map are derived from the same geometric description of the maze (the world file and its wall mesh), so no separate mapping run is required when the maze geometry is known.

\subsection{Planning}

The planner runs online. Whenever a goal arrives, it reads the current pose from the \texttt{map}$\to$\texttt{base\_footprint} transform and plans from there, so the same node serves the initial goal and any later goal sent by an operator. It does not replan otherwise; a new plan replaces the previous one in the controller.

\subsubsection{Occupancy grid}
The walls of the maze are rasterized into a grid with resolution $\delta = 0.1$~m. A cell is blocked if its center lies within an inflation radius $r_{\text{inf}} = 0.6$~m of any wall segment, so the center of every free cell is at least $r_{\text{inf}}$ from every wall, any point inside a free cell at least $r_{\text{inf}} - \delta/\sqrt{2} \approx 0.53$~m, and the planners can treat the robot as a point. The inflation must satisfy $r_{\text{inf}} \geq \delta/\sqrt{2}$ so that no wall slips between cell centers.

\subsubsection{A* with shortcutting}
The default planner runs 8-connected A* over free cells with the octile distance as heuristic. Diagonal moves are not allowed to cut a blocked corner. The resulting path follows the grid directions, so it is post-processed with line-of-sight shortcutting (Algorithm~\ref{alg:shortcut}): starting from the first point, the farthest later point that is still visible through free cells is kept, and the process repeats from there. Visibility is checked by sampling the segment every $\delta/2$. Each remaining point is given the heading of the segment that leaves it.

\begin{algorithm}
\caption{Line-of-Sight Shortcutting}\label{alg:shortcut}
\begin{algorithmic}
\STATE {\textsc{Shortcut}}$(\mathcal{G}, \mathcal{P} = \{\mathbf{p}_1, \ldots, \mathbf{p}_n\})$
\STATE $\mathcal{S} \gets \{\mathbf{p}_1\}$, $i \gets 1$
\WHILE{$i < n$}
    \STATE $j \gets i + 1$
    \WHILE{$j < n$ \AND \textsc{Visible}$(\mathcal{G}, \mathbf{p}_i, \mathbf{p}_{j+1})$}
        \STATE $j \gets j + 1$
    \ENDWHILE
    \STATE $\mathcal{S} \gets \mathcal{S} \cup \{\mathbf{p}_j\}$, $i \gets j$
\ENDWHILE
\RETURN $\mathcal{S}$
\end{algorithmic}
\end{algorithm}

\subsubsection{Hybrid A*}
As an alternative, Hybrid A* \cite{dolgov08gppSTAIR} searches over continuous poses $[x, y, \theta]^T$. Each expansion drives forward along one of five arcs of length $0.3$~m whose curvatures are evenly spaced in $[-1/r_{\min}, 1/r_{\min}]$ with $r_{\min} = 1$~m. Only forward motion is generated because the controller does not reverse. States that fall in the same grid cell and in the same of 72 heading bins are merged. Each arc costs its length times $(1 + c_\kappa |\kappa| r_{\min})$, with $c_\kappa = 0.1$, which favors straight motion. The heuristic is the maximum of the Euclidean distance to the goal and the grid path length to the goal around walls, which is precomputed with a single Dijkstra search from the goal. In a maze the Euclidean distance alone is a poor estimate: without the obstacle-aware term, the search does not reach the goal of our experiments within $2 \times 10^6$ expansions (Table~\ref{tab:planners}). The search ends at the first expanded state within $0.3$~m of the goal.

\subsubsection{Reference generation}
The controller densifies the received path into a reference with spacing of at most $v_{\text{ref}} \Delta t$ (each segment is split into equal steps), where $v_{\text{ref}} = 0.5$~m/s is the reference speed and $\Delta t = 0.1$~s the controller sampling time, so consecutive reference points are about one control step apart at the desired speed. Only the path's positions are used; each reference heading is that of its segment. At each step, a path tracker finds the reference point closest to the robot within a forward window of 200 points from its previous index (the whole path on the first step), so the index never moves backward; this prevents it from jumping to a later part of the path that passes nearby on the other side of a wall. The NMPC receives the reference from that point onward.

\subsection{Mapping}

\begin{figure}[!t]
\centering
\includegraphics[width=\linewidth]{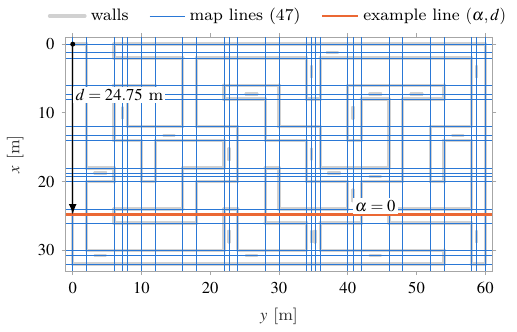}
\caption{Line map of the maze: the 47 infinite lines (blue) extracted from the vertical faces of the maze mesh (gray). Short end faces are skipped, and collinear faces of different walls share one line. One line (orange) is shown with its Hesse parameters: the normal angle $\alpha$ and the distance $d$ from the world origin $O$.}
\label{fig:hesse-map}
\end{figure}

The localization layer represents the maze as a set of infinite lines in Hesse normal form: each wall face is a line $\{\mathbf{p} : \hat{\mathbf{n}}(\alpha)^T \mathbf{p} = d\}$ with normal angle $\alpha$ and distance $d$ to the origin of the world frame, where $\hat{\mathbf{n}}(\alpha) = [\cos \alpha, \sin \alpha]^T$. The map $\mathcal{M}$ is the set of pairs $(\alpha_m, d_m)$ (Fig.~\ref{fig:hesse-map}).

The lines are extracted directly from the geometry of the simulated world. For a maze given as a triangle mesh, each triangle with normal $\mathbf{n}$ is kept only if it is a vertical face ($|n_z| / \|\mathbf{n}\| \leq 0.1$) and its horizontal extent is at least $0.5$~m, which discards floors, wall tops, and short end faces. The kept triangle yields the line with $\alpha = \mathrm{atan2}(n_y, n_x)$ and $d = \hat{\mathbf{n}}^T \mathbf{p}_0$ for any vertex $\mathbf{p}_0$. Walls given as boxes contribute their four faces in the same way. Lines are stored in a canonical form with $d \geq 0$, and duplicates (the many coplanar triangles of a wall face) are merged.

A wall face is visible only from one side, but the canonical form does not say which. At run time, each map line is therefore oriented so that its normal points from the robot toward the wall: if the signed distance from the predicted robot position, $d_m - \hat{\mathbf{n}}(\alpha_m)^T [\hat{x}, \hat{y}]^T$, is negative, the line is used as $(\alpha_m + \pi, -d_m)$. This matches the convention of the observed lines, whose distance to the robot is always positive.

\subsection{Localization}

The robot's state is $\mathbf{x} = [x, y, \theta]^T$, its planar pose in the map frame. An EKF predicts the state with wheel odometry and corrects it with lines extracted from the LiDAR scan.

An earlier prototype of this stack also implemented a range--bearing measurement model with point landmarks at known positions. It is simple and accurate, but it requires installing beacons in the environment and keeping their surveyed positions up to date. Line features need no infrastructure, since the walls themselves are the landmarks, so the current stack uses only the line model.

\subsubsection{Prediction}
The odometry source reports a pose at a high rate. Between two consecutive odometry poses, the heading increment is $\Delta \beta = \mathrm{wrap}(\theta^o_k - \theta^o_{k-1})$ and the chord is $c = \|[\Delta x^o, \Delta y^o]\|$. Assuming the robot moved along a circular arc, the traveled distance is
\begin{equation} \label{eq:arc-length}
    \Delta d = \pm\, c \, \frac{|\Delta \beta|}{2 \sin(|\Delta \beta| / 2)},
\end{equation}
with $\Delta d = c$ when $|\Delta \beta|$ is negligible, and with the sign negative when the chord points backward with respect to the mid-point heading. The state is propagated with the mid-point model:
\begin{subequations}
\begin{align} \label{eq:odometry-model}
    \hat{\mathbf{x}}_{k+1|k} &= f(\hat{\mathbf{x}}_{k|k}, \mathbf{u}_k) \\
    &= \begin{bmatrix} \hat{x}_{k|k} + \Delta d \cos \theta_{m} \\ \hat{y}_{k|k} + \Delta d \sin \theta_{m} \\ \hat{\theta}_{k|k} + \Delta \beta \end{bmatrix}, \quad \theta_m = \hat{\theta}_{k|k} + \frac{\Delta \beta}{2},
\end{align}
\end{subequations}
where $\mathbf{u}_k = [\Delta d, \Delta \beta]^T$.

Propagating only the noise of $\mathbf{u}_k$ through the control Jacobian $W_k$ underestimates the uncertainty of a skid-steered or mecanum platform, whose wheels slip sideways and in rotation. The process noise therefore adds two terms that grow with the motion:
\begin{equation} \label{eq:process-noise}
    Q_k = W_k \Sigma_u W_k^T + \begin{bmatrix} \sigma_{\perp}^2 |\Delta d| \, \mathbf{n}_\perp \mathbf{n}_\perp^T & \mathbf{0} \\ \mathbf{0}^T & \sigma_{\theta}^2 (|\Delta d| + |\Delta \beta|) \end{bmatrix},
\end{equation}
where $\Sigma_u = \mathrm{diag}(\sigma_{\Delta d}^2, \sigma_{\Delta \beta}^2)$ is the per-step odometry noise, $\mathbf{n}_\perp = [-\sin \theta_m, \cos \theta_m]^T$ is the lateral direction, and $\sigma_\perp$ and $\sigma_\theta$ are the lateral and heading slip densities. We use $\sigma_{\Delta d} = 0.002$~m and $\sigma_{\Delta \beta} = 0.002$~rad per odometry message (50~Hz), $\sigma_\perp = 0.1$~m$/\sqrt{\text{m}}$, and $\sigma_\theta = 0.1$. Algorithm~\ref{alg:ekf-prediction} summarizes the prediction.

\begin{algorithm}[!t]
\caption{EKF Prediction Step}\label{alg:ekf-prediction}
\begin{algorithmic}
\STATE {\textsc{Predict}}$(\hat{\mathbf{x}}_{k|k}, P_{k|k}, \Delta d, \Delta \beta)$
\STATE $\theta_{m} \gets \hat{\theta}_{k|k} + \frac{\Delta \beta}{2}$ \COMMENT{Mid-point heading}
\STATE $\hat{\mathbf{x}}_{k+1|k} \gets f(\hat{\mathbf{x}}_{k|k}, \mathbf{u}_k)$
\STATE $\hat{\theta}_{k+1|k} \gets \text{wrap}_{[-\pi,\pi]}(\hat{\theta}_{k+1|k})$
\STATE $A_k \gets \nabla_\mathbf{x} f(\hat{\mathbf{x}}_{k|k}, \mathbf{u}_k)$, $W_k \gets \nabla_\mathbf{u} f(\hat{\mathbf{x}}_{k|k}, \mathbf{u}_k)$
\STATE $Q_k \gets$ Eq.~\eqref{eq:process-noise}
\STATE $P_{k+1|k} \gets A_k P_{k|k} A_k^T + Q_k$
\RETURN $\hat{\mathbf{x}}_{k+1|k}, P_{k+1|k}$
\end{algorithmic}
\end{algorithm}

\subsubsection{Scan preprocessing}
Scans arrive at 10~Hz and are stamped at a different time than the last odometry message, to which the EKF state refers. Before line extraction, the whole scan is expressed in the robot frame at the EKF state's time by chaining the LiDAR pose at the scan stamp through the odometry frame. Motion within a single scan is not compensated. This removes the apparent rotation of the walls during fast turns.

\subsubsection{Line extraction}
Lines are extracted from the scan points (in the robot frame) with RANSAC \cite{fischler1981ransac}, as shown in Algorithm~\ref{alg:line-fitting}. In each round, $n_{\text{iter}} = 30$ pairs of points are sampled; pairs closer than $\epsilon = 5$~cm are discarded, and each remaining pair defines a candidate line, and points closer than $\gamma = 3$~cm to it are its inliers. The candidate with the most inliers is kept if it has at least $n_{\min} = 10$ of them, its inliers are removed, and the next round starts. A fixed random seed makes the extraction deterministic.

\begin{algorithm}
\caption{RANSAC-based Line Extraction}\label{alg:line-fitting}
\begin{algorithmic}
\STATE {\textsc{FitLines}}$(\mathcal{P}, \gamma, n_{\min}, n_{\text{iter}})$
\STATE $\mathcal{L} \gets \emptyset$ \COMMENT{Detected lines}
\WHILE{$|\mathcal{P}| > n_{\min}$}
    \STATE $\mathcal{I}_{best} \gets \emptyset$
    \FOR{$i = 1$ to $n_{\text{iter}}$}
        \STATE $\mathbf{p}_1, \mathbf{p}_2 \gets \textsc{RandomSample}(\mathcal{P}, 2)$
        \IF{$\|\mathbf{p}_2 - \mathbf{p}_1\| < \epsilon$}
            \STATE \textbf{continue} \COMMENT{Points too close}
        \ENDIF
        \STATE $\Delta \mathbf{p} \gets \mathbf{p}_2 - \mathbf{p}_1$
        \STATE $\hat{\mathbf{n}} \gets [-\Delta p_y, \Delta p_x]^T / \|\Delta \mathbf{p}\|$ \COMMENT{Unit normal}
        \STATE $d \gets \hat{\mathbf{n}}^T \mathbf{p}_1$
        \STATE $\mathcal{I}_i \gets \{\mathbf{p} \in \mathcal{P} : |\hat{\mathbf{n}}^T \mathbf{p} - d| < \gamma\}$ \COMMENT{Inliers}
        \IF{$|\mathcal{I}_i| > |\mathcal{I}_{best}|$}
            \STATE $\mathcal{I}_{best} \gets \mathcal{I}_i$
        \ENDIF
    \ENDFOR
    \IF{$|\mathcal{I}_{best}| < n_{\min}$}
        \STATE \textbf{break} \COMMENT{No more lines to extract}
    \ENDIF
    \STATE $\mathcal{L} \gets \mathcal{L} \cup \{\textsc{TotalLeastSquares}(\mathcal{I}_{best})\}$
    \STATE $\mathcal{P} \gets \mathcal{P} \setminus \mathcal{I}_{best}$ \COMMENT{Remove used points}
\ENDWHILE
\RETURN $\mathcal{L}$
\end{algorithmic}
\end{algorithm}

Each set of inliers is refined with total least squares (TLS), shown in Algorithm~\ref{alg:tls}, which minimizes perpendicular distances from the points to the line rather than vertical distances. The normal $\hat{\mathbf{n}}$ is the eigenvector of the inliers' scatter matrix with the smallest eigenvalue, i.e., the direction of least spread. The uncertainty of the line comes from the residuals $r_i = \hat{\mathbf{n}}^T \mathbf{p}_i - d$: their sample standard deviation $\tilde{\sigma}_d$ gives $\sigma_d$, and $\tilde{\sigma}_d$ divided by the half-length $l$ of the segment along its tangent $\boldsymbol{\tau} = [-\hat{n}_y, \hat{n}_x]^T$ gives $\sigma_\alpha$:
\begin{subequations}
\begin{align}
    l &= \max_{i} \left|\boldsymbol{\tau}^T(\mathbf{p}_{i} - \boldsymbol{\mu})\right|, \label{eq:line-extent} \\
    \tilde{\sigma}_d &= \sqrt{\tfrac{1}{n-1} \textstyle\sum_{i=1}^{n} (r_i - \bar{r})^2}, \quad \sigma_d = \max(\tilde{\sigma}_d, \sigma_{d,\min}), \label{eq:sigma-d} \\
    \sigma_{\alpha} &= \max\!\left(\frac{\tilde{\sigma}_d}{\max(l, l_{\min})},\ \sigma_{\alpha,\min}\right). \label{eq:sigma-alpha}
\end{align}
\end{subequations}
The floors $\sigma_{d,\min} = 2$~cm, $\sigma_{\alpha,\min} = 1^\circ$, and $l_{\min} = 0.1$~m prevent a short or nearly perfect fit from claiming an unrealistically small uncertainty. Finally, the line is put in canonical form with $d \geq 0$, so its normal points from the robot toward the wall, and the extent $[s_{\min}, s_{\max}]$ of its inliers along the tangent is recorded for SLAM.

\begin{algorithm}
\caption{Total Least Squares Line Fitting}\label{alg:tls}
\begin{algorithmic}
\STATE {\textsc{TotalLeastSquares}}$(\mathcal{I})$ \COMMENT{Inlier point set, $n = |\mathcal{I}|$}
\STATE $\boldsymbol{\mu} \gets \frac{1}{n} \sum_{i} \mathbf{p}_i$ \COMMENT{Mean}
\STATE $C \gets \sum_{i} (\mathbf{p}_i - \boldsymbol{\mu})(\mathbf{p}_i - \boldsymbol{\mu})^T$ \COMMENT{Scatter matrix}
\STATE $\hat{\mathbf{n}} \gets$ eigenvector of $C$ with the smallest eigenvalue
\STATE $\boldsymbol{\tau} \gets [-\hat{n}_y, \hat{n}_x]^T$ \COMMENT{Line direction}
\STATE $d \gets \hat{\mathbf{n}}^T \boldsymbol{\mu}$, $\alpha \gets \mathrm{atan2}(\hat{n}_y, \hat{n}_x)$
\STATE $\sigma_d, \sigma_\alpha \gets$ Eqs.~\eqref{eq:line-extent}--\eqref{eq:sigma-alpha}
\IF{$d < 0$}
    \STATE $d \gets -d$, $\alpha \gets \text{wrap}(\alpha + \pi)$ \COMMENT{Canonical form}
\ENDIF
\RETURN $(\alpha, d, \sigma_\alpha, \sigma_d)$
\end{algorithmic}
\end{algorithm}

\subsubsection{Measurement model}
An observed line $\mathbf{z}_i = [\alpha_i, d_i]^T$ is expressed in the robot frame. For a map line $(\alpha_m, d_m)$, oriented toward the robot as described above, the predicted observation and its Jacobian are
\begin{equation} \label{eq:ekf-lines-meas-func}
    h(\hat{\mathbf{x}}) = \begin{bmatrix}
        \alpha_{m} - \hat{\theta} \\
        d_{m} - \hat{x} \cos \alpha_m - \hat{y} \sin \alpha_m
    \end{bmatrix},
\end{equation}
\begin{equation} \label{eq:ekf-lines-jacobian}
    H = \begin{bmatrix}
        0 & 0 & -1 \\
        -\cos{\alpha_{m}} & -\sin{\alpha_{m}} & 0
    \end{bmatrix},
\end{equation}
with measurement noise $R_i = \mathrm{diag}(\sigma_{\alpha,i}^2, \sigma_{d,i}^2)$ from the line fit. Throughout the paper, line observations, innovations, and their covariances are ordered as $[\alpha, d]$.

\subsubsection{Data association and update}
Each observed line is associated with the map line that minimizes the score $w_\alpha |\Delta \alpha| + w_d |\Delta d|$, where $\Delta \alpha$ is the difference between the map line's angle and the observed angle rotated into the world frame, $\Delta d$ is the difference between the observed distance and the predicted one, $w_\alpha = 2$, and $w_d = 1$. Only the best-scoring map line is considered: if $|\Delta \alpha| \geq 30^\circ$ or $|\Delta d| \geq 0.5$~m, the observation is discarded rather than matched to the next candidate. An accepted pair must also pass a $\chi^2$ test on the Mahalanobis distance of its innovation, with threshold $t_{\chi^2} = 9.21$ (the 99\% quantile for two degrees of freedom). The updates are applied sequentially, one line at a time, as shown in Algorithm~\ref{alg:ekf-update-lines}.

\begin{algorithm}
\caption{EKF Line-based Update}\label{alg:ekf-update-lines}
\begin{algorithmic}
\STATE {\textsc{LinesUpdateEKF}}$(\hat{\mathbf{x}}, P, \mathcal{Z}, \mathcal{M}, t_{\chi^2})$
\FOR{each line $\mathbf{z}_i = (\alpha_i, d_i, \sigma_{\alpha,i}, \sigma_{d,i}) \in \mathcal{Z}$}
    \STATE $j \gets \textsc{AssociateLine}(\mathbf{z}_i, \hat{\mathbf{x}}, \mathcal{M})$
    \IF{$j = $ \text{NULL}}
        \STATE \textbf{continue} \COMMENT{No association found}
    \ENDIF
    \STATE $(\alpha_m, d_m) \gets \textsc{Oriented}(\mathcal{M}_j, \hat{\mathbf{x}})$
    \STATE $R_i \gets \text{diag}(\sigma^2_{\alpha,i}, \sigma^2_{d,i})$
    \STATE $[\hat{\alpha}_i, \hat{d}_i]^T \gets h(\hat{\mathbf{x}})$ \COMMENT{Eq.~\eqref{eq:ekf-lines-meas-func}}
    \STATE $\boldsymbol{\nu}_i \gets [\text{wrap}(\alpha_i - \hat{\alpha}_i),\ d_i - \hat{d}_i]^T$ \COMMENT{Innovation}
    \STATE $H \gets$ Eq.~\eqref{eq:ekf-lines-jacobian}
    \STATE $S_i \gets H P H^T + R_i$ \COMMENT{Innovation covariance}
    \IF{$\boldsymbol{\nu}_i^T S_i^{-1} \boldsymbol{\nu}_i > t_{\chi^2}$}
        \STATE \textbf{continue} \COMMENT{Reject outlier}
    \ENDIF
    \STATE $K_i \gets P H^T S_i^{-1}$ \COMMENT{Kalman gain}
    \STATE $\hat{\mathbf{x}} \gets \hat{\mathbf{x}} + K_i \boldsymbol{\nu}_i$, $\hat{\theta} \gets \text{wrap}(\hat{\theta})$
    \STATE $P \gets (I - K_i H) P$
\ENDFOR
\RETURN $\hat{\mathbf{x}}, P$
\end{algorithmic}
\end{algorithm}

The localization node publishes the estimate as the \texttt{map}$\to$\texttt{odom} transform, following REP~105, so that \texttt{map}$\to$\texttt{odom}$\to$\texttt{base\_footprint} equals the EKF pose while the odometry frame remains continuous. The robot frame \texttt{base\_footprint} is the ground projection of \texttt{base\_link}, following the convention of REP~120.

\subsection{EKF SLAM with Line Landmarks}

When no geometric description of the maze is available, the same front end builds the line map online. The state is augmented with the lines mapped so far,
\begin{equation}
    \mathbf{x} = [x, y, \theta, \alpha_1, d_1, \ldots, \alpha_n, d_n]^T \in \mathbb{R}^{3 + 2n},
\end{equation}
with joint covariance $P$. The robot's start pose is taken as known, with zero covariance (by default it defines the map frame); this anchors the map and removes the gauge freedom of SLAM.

\subsubsection{Prediction}
The map is static, so only the pose evolves with Eq.~\eqref{eq:odometry-model}. With $P$ partitioned into pose ($\mathbf{x}\mathbf{x}$) and map ($\mathbf{x}\mathcal{M}$, $\mathcal{M}\mathcal{M}$) blocks, the prediction is
\begin{equation}
    P_{\mathbf{x}\mathbf{x}} \gets A_k P_{\mathbf{x}\mathbf{x}} A_k^T + Q_k, \qquad P_{\mathbf{x}\mathcal{M}} \gets A_k P_{\mathbf{x}\mathcal{M}},
\end{equation}
which costs $O(n)$ instead of $O(n^2)$.

\subsubsection{Measurement model}
The predicted observation of landmark $j$ is Eq.~\eqref{eq:ekf-lines-meas-func} with $(\alpha_m, d_m) = (\alpha_j, d_j)$, but now the landmark is part of the state, so the Jacobian has a pose block and a landmark block:
\begin{equation}
    H_{\mathbf{x}} = \begin{bmatrix} 0 & 0 & -1 \\ -c_j & -s_j & 0 \end{bmatrix}, \quad
    H_{j} = \begin{bmatrix} 1 & 0 \\ s_j \hat{x} - c_j \hat{y} & 1 \end{bmatrix},
\end{equation}
where $c_j = \cos \alpha_j$ and $s_j = \sin \alpha_j$. The full Jacobian is zero elsewhere, so the innovation covariance $S$ only needs the pose block, the landmark block, and their cross-covariance.

\subsubsection{Landmark initialization}
A new line is initialized by inverting the measurement model. From an observation $(\alpha_z, d_z)$ at pose $(x, y, \theta)$,
\begin{equation}
    \alpha = \alpha_z + \theta, \qquad d = d_z + x \cos \alpha + y \sin \alpha,
\end{equation}
with Jacobians with respect to the pose and the observation
\begin{equation}
    G_{\mathbf{x}} = \begin{bmatrix} 0 & 0 & 1 \\ \cos \alpha & \sin \alpha & e \end{bmatrix}, \quad
    G_{\mathbf{z}} = \begin{bmatrix} 1 & 0 \\ e & 1 \end{bmatrix},
\end{equation}
where $e = -x \sin \alpha + y \cos \alpha$. The state and covariance are augmented with
\begin{subequations}
\begin{align}
    P_{\mathcal{L}\mathcal{L}} &= G_{\mathbf{x}} P_{\mathbf{x}\mathbf{x}} G_{\mathbf{x}}^T + G_{\mathbf{z}} R \, G_{\mathbf{z}}^T, \\
    P_{\mathcal{L}\cdot} &= G_{\mathbf{x}} P_{\mathbf{x}\cdot},
\end{align}
\end{subequations}
where $P_{\mathbf{x}\cdot}$ denotes the pose rows of $P$, so the new line is correlated with the pose and with every existing landmark.

\subsubsection{Data association}
Infinite lines are ambiguous in a maze: two collinear walls in different corridors, or the two faces of a thin wall, have nearly the same parameters. Each landmark therefore also keeps the extent $[s_{\min}, s_{\max}]$ it has been seen over along its tangent, and a landmark is a candidate match for an observation only if (i) the robot is on the side the wall was seen from (predicted $\hat{d} > 0$), and (ii) the observed segment, projected onto the landmark's tangent, overlaps its extent within a margin of $0.5$~m. Among the candidates, the one with the smallest Mahalanobis distance $m^2$ is chosen. Three outcomes follow:
\begin{itemize}
    \item $m^2 < 9.21$: the observation updates the state, and the landmark's extent grows to include the observed segment.
    \item $9.21 \leq m^2 < t_{\text{new}} = 50$: the observation is ambiguous, close to a mapped line but not close enough, and is dropped.
    \item $m^2 \geq t_{\text{new}}$, or no candidate: the observation may be a new wall.
\end{itemize}
The middle band keeps a slightly misfit observation of a known wall from creating a duplicate landmark.

\subsubsection{Landmark promotion}
A single bad line fit should never enter the map, since a landmark cannot be removed once its correlations are in $P$. A possible new wall therefore first becomes a candidate, kept outside the state. Later observations within $5^\circ$ and $0.1$~m of a candidate in the world frame, with an overlapping extent, count as re-sightings. A candidate seen $n_{\text{obs}} = 3$ times is promoted to a landmark, initialized as above from its latest sighting, with the extent accumulated over all its sightings; a candidate not seen for more than $n_{\text{timeout}} = 10$ scans is discarded. Algorithm~\ref{alg:ekf-slam-update} summarizes the update.

\begin{algorithm}
\caption{EKF SLAM Line Update}\label{alg:ekf-slam-update}
\begin{algorithmic}
\STATE {\textsc{SlamUpdate}}$(\hat{\mathbf{x}}, P, \mathcal{Z}, \mathcal{C})$ \COMMENT{$\mathcal{C}$: candidates}
\FOR{each line $\mathbf{z}_i \in \mathcal{Z}$}
    \STATE $(j, m^2) \gets \arg\min$ Mahalanobis over landmarks seen from the same side whose extent overlaps $\mathbf{z}_i$
    \IF{$j \neq$ NULL \AND $m^2 < t_{\chi^2}$}
        \STATE EKF update of $(\hat{\mathbf{x}}, P)$ with $H = [H_{\mathbf{x}}, \ldots, H_j, \ldots]$
        \STATE Grow extent of landmark $j$
    \ELSIF{$j =$ NULL \OR $m^2 \geq t_{\text{new}}$}
        \STATE Add $\mathbf{z}_i$ to its candidate in $\mathcal{C}$ (or a new one)
        \IF{the candidate has $n_{\text{obs}}$ sightings}
            \STATE Augment $(\hat{\mathbf{x}}, P)$ with the line; remove it from $\mathcal{C}$
        \ENDIF
    \ELSE
        \STATE Drop $\mathbf{z}_i$ \COMMENT{Ambiguous}
    \ENDIF
\ENDFOR
\STATE Discard candidates unseen for more than $n_{\text{timeout}}$ scans
\RETURN $\hat{\mathbf{x}}, P, \mathcal{C}$
\end{algorithmic}
\end{algorithm}

\subsection{Control}

The controller is a nonlinear MPC that tracks the reference while discrete-time CBF constraints keep the predicted trajectory away from the LiDAR returns.

\subsubsection{Model and cost}
The robot is controlled as a unicycle with input $\mathbf{u} = [v, \omega]^T$, discretized with forward Euler at $\Delta t = 0.1$~s:
\begin{equation}
    \mathbf{x}_{k+1} = \mathbf{x}_k + \Delta t \begin{bmatrix} v_k \cos \theta_k & v_k \sin \theta_k & \omega_k \end{bmatrix}^T.
\end{equation}
Over a horizon of $N = 20$ steps (2~s), the NMPC solves
\begin{subequations} \label{eq:nmpc}
\begin{align}
    \min_{\mathbf{u}_{0:N-1}, \mathbf{x}_{0:N}, \mathbf{s}} \ & \sum_{k=0}^{N} \|\mathbf{x}_k - \mathbf{x}^r_k\|^2_Q + \sum_{k=0}^{N-1} \Big( \|\mathbf{u}_k\|^2_R \notag \\
    & + \|\mathbf{u}_k - \mathbf{u}_{k-1}\|^2_{R_\Delta} + \rho \sum_{i=1}^{M} s_{i,k} \Big) \label{eq:cost-function} \\
    \text{s.t.} \quad & \mathbf{x}_0 = \hat{\mathbf{x}}, \quad \mathbf{x}_{k+1} = f(\mathbf{x}_k, \mathbf{u}_k), \\
    & v_{\min} \leq v_k \leq v_{\max}, \quad |\omega_k| \leq \omega_{\max}, \\
    & h_i(\mathbf{x}_{k+1}) - (1 - \gamma) h_i(\mathbf{x}_k) + s_{i,k} \geq 0, \label{eq:dcbf} \\
    & s_{i,k} \geq 0, \quad i = 1, \ldots, M, \ k = 0, \ldots, N-1,
\end{align}
\end{subequations}
where $\mathbf{x}^r_k$ is the reference, whose headings are unwrapped to be continuous from the current heading, and $\mathbf{u}_{-1}$ is the command sent at the previous step (zero after a new plan is received). The rate term $R_\Delta$ penalizes command changes, including against the command already sent; without it, the solver alternates between nearly equal plans that pass an obstacle on the left and on the right. We use $Q = \mathrm{diag}(1000, 1000, 100)$, $R = I$, $R_\Delta = 10 I$, $v \in [0, 1]$~m/s, and $|\omega| \leq 2$~rad/s; the lower speed bound is zero because the robot only drives forward.

\subsubsection{Discrete-time CBF}
Each obstacle point $\mathbf{o}_i$ defines the barrier
\begin{equation} \label{eq:h-func}
    h_i(\mathbf{x}) = \|[x, y]^T - \mathbf{o}_i\|^2 - d_{\text{safe}}^2,
\end{equation}
which is nonnegative when the robot's center is at least $d_{\text{safe}} = 0.5$~m from the point. Constraint~\eqref{eq:dcbf} is the discrete-time CBF condition of Zeng et al. \cite{9483029}: along the predicted trajectory, $h_i$ may decrease by at most a fraction $\gamma \in (0, 1]$ per step, so the robot can approach an obstacle but only at a geometrically decaying rate. A smaller $\gamma$ makes the robot react earlier; we use $\gamma = 0.1$. Because the condition is written directly on the discrete predicted states of the MPC, it needs no time derivatives of $h_i$, unlike continuous-time and high-order CBFs \cite{9516971}, and the optimizer can steer, not only brake, to keep $h_i$ positive. The slack variables $s_{i,k}$, penalized linearly with $\rho = 10^4$, keep the problem feasible when the robot is boxed in; with a linear (exact) penalty, the slack stays zero whenever the constraint can be satisfied, provided $\rho$ exceeds the constraint's Lagrange multiplier.

\subsubsection{Obstacle selection}
The LiDAR points are transformed into the map frame with the robot pose at the scan time, so they stay fixed in the world however much the robot has moved since the scan. Every point is a potential obstacle, but a scan has thousands of points and each one adds $N$ constraints. Points farther than $3$~m or outside $\pm 120^\circ$ of the heading are ignored, and the remaining field of view is divided into $M = 16$ equal sectors; the nearest point in each sector becomes an obstacle. Sectors ensure that a long wall cannot take every slot and hide a smaller obstacle. Unused slots are filled with a point far from the robot so that the problem keeps a fixed size and is built only once.

\subsubsection{Real-time implementation}
The problem is built once with CasADi \cite{Andersson2019}, expanded into scalar symbolic form for fast evaluation of derivatives, and solved with IPOPT \cite{wachter2006ipopt}. The solver's primal variables are initialized with the previous plan shifted by one step and the states it produces from the current pose; a guess far from the robot, such as all zeros, makes IPOPT fail because the barriers depend on the states. The solver's CPU time is capped at $0.8 \Delta t$, since a late command is worse than a failed solve. If a solve fails, the controller keeps executing the previous plan, shifted, for up to 5 steps, and then stops. The controller also stops if the pose is older than $0.5$~s or the latest scan older than $0.3$~s, which happens if localization or the LiDAR fails.

\section{Platform and Implementation}

\subsection{Simulated robot}
The robot is a Husarion ROSbot XL simulated in Gazebo \cite{koenig2004gazebo}. Its four mecanum wheels are driven as a differential drive by the ros2\_control differential drive controller, with wheel separation $0.44$~m and wheel radius $0.05$~m. Gazebo approximates mecanum rollers with anisotropic friction, under which the nominal kinematics overestimate distance and underestimate rotation; the controller's wheel radius and separation multipliers ($0.907$ and $0.85$) were tuned to reduce the average odometry error. The error that remains depends on the motion: in sharp turns the wheels slide, and odometry misses about 20\% of the rotation in a typical turn and up to 40\% in the sharpest ones, which no constant multiplier can correct (Section~\ref{sec:experiments}). Forward acceleration is limited to $1$~m/s$^2$ to avoid wheel slip on takeoff, which corrupts the odometry heading. The turn rate is not limited, since the NMPC model assumes it changes instantly and lag makes the closed loop oscillate.

The robot carries a simulated Slamtec RPLIDAR S3: 3200 points over $360^\circ$ ($0.1125^\circ$ apart) at 10~Hz, with a range of $0.05$ to $40$~m and Gaussian range noise with $\sigma = 15$~mm, taken from the datasheet accuracy of $\pm 30$~mm as roughly $2\sigma$.

\subsection{Environment}
\begin{figure}[!t]
\centering
\includegraphics[width=\linewidth]{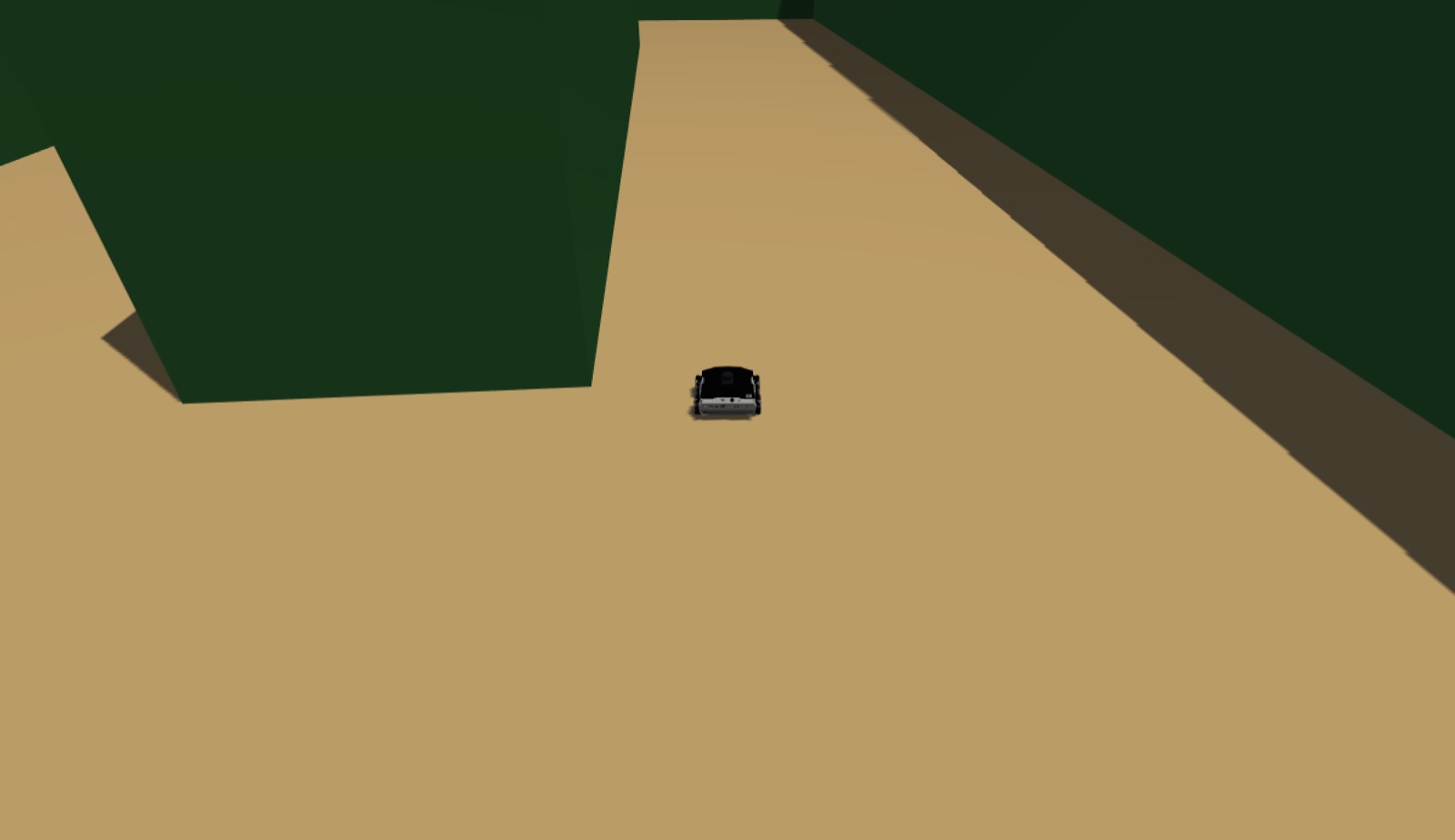}\\[2pt]
{\footnotesize (a)}\\[4pt]
\includegraphics[width=\linewidth]{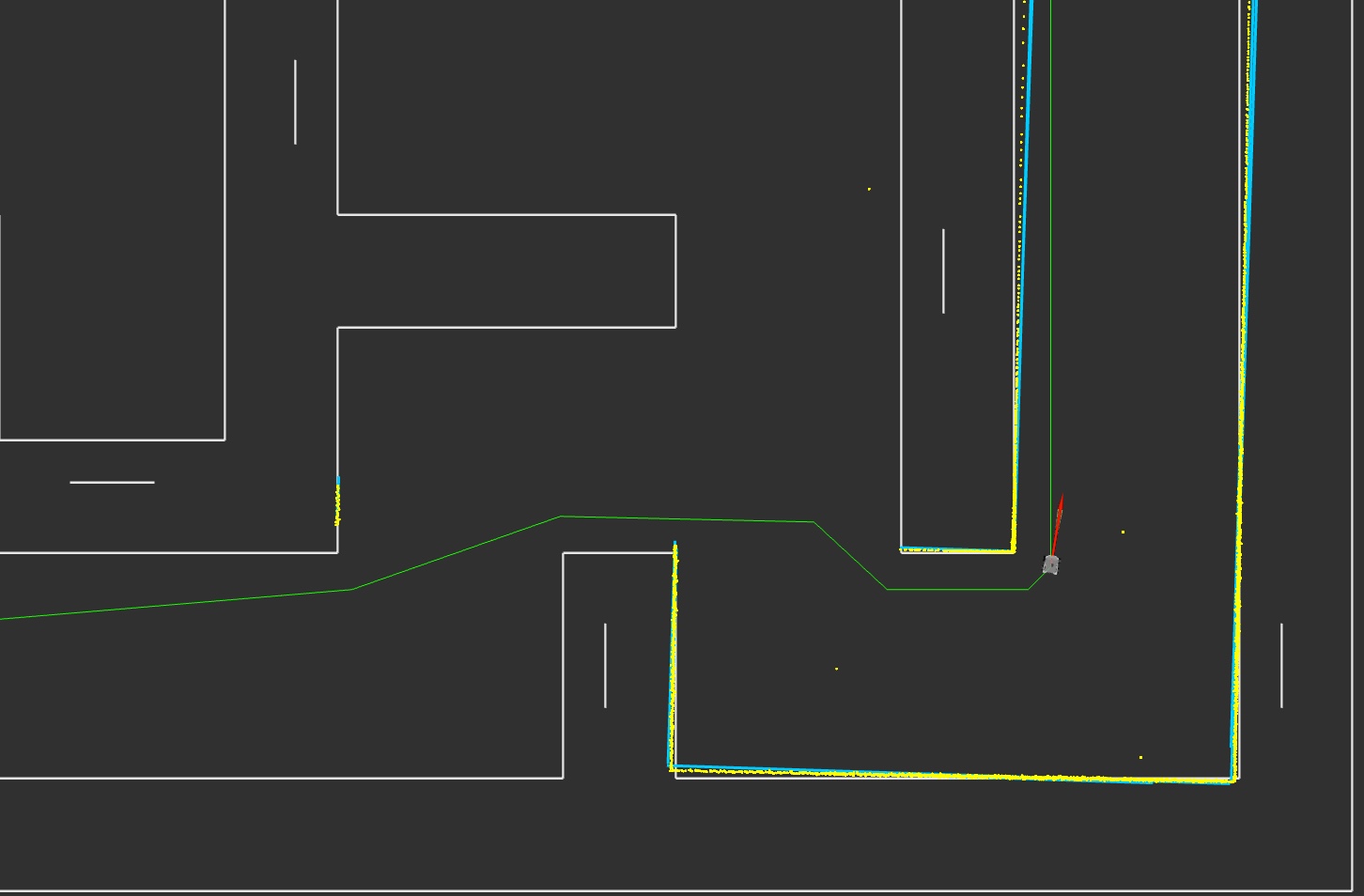}\\[2pt]
{\footnotesize (b)}
\caption{Simulation environment, at the same instant during a run. (a)~The ROSbot XL entering a turn in the maze, imported from STL in Gazebo. (b)~The same moment in RViz, top-down: map walls (white), LiDAR scan (yellow), lines extracted from the scan (cyan), NMPC reference path (green), and EKF pose (red arrow).}
\label{fig:gazebo-maze}
\end{figure}

The maze is the same model used in the earlier prototype, designed in Autodesk Inventor and imported into Gazebo as an STL mesh with collision (Fig.~\ref{fig:gazebo-maze}). The line map for localization is extracted from this same mesh, as described in the Mapping section, and the planner's occupancy grid from the same walls. Gazebo also publishes the robot's ground-truth pose, which is used only for evaluation.

\subsection{Software}
The stack targets ROS~2 Lyrical \cite{macenski2022ros2} and Gazebo Jetty and is written entirely in C++. It is split into four packages: \texttt{maze\_runner\_core} holds the algorithms (line extraction, map lines, EKF, EKF SLAM, occupancy grid, planners, NMPC, path tracker) with no ROS dependency, together with their unit tests; \texttt{maze\_runner\_nodes} wraps them in three ROS~2 nodes, one per layer; and two further packages hold the robot description and the simulation worlds and launch files. Within the core, free functions are pure, and only components with state, such as the filters and the controller, are classes. All dependencies, including ROS~2, Gazebo, and CasADi, are installed from RoboStack through the pixi package manager and pinned in a lock file, so building, testing, and running the stack each take a single command on Linux and macOS. Every run can record a bag with the ground-truth and estimated poses, wheel odometry, commands, scans, the plan, the NMPC reference, and the NMPC solve times, from which all results in this paper are computed. The scripts that turn the bags into every figure and number in this paper are in the repository under \texttt{.paper/scripts}.

\section{Experimental Validation}\label{sec:experiments}

All experiments run in Gazebo with the full stack in closed loop. Unless stated otherwise, the robot starts at $(2.39, 4.31)$~m with heading $0.045$~rad near the maze entrance and is sent to a goal at $(29.83, 55.90)$~m near the exit. Each result comes from a single run on an Apple M5 Pro laptop, with the Gazebo GUI and RViz open; the simulation ran at a real-time factor of about 1.0. The planner (A* with shortcutting) produced a 26-point path in 6.3~ms, and the robot drove the resulting 135~m route in 269~s, stopping 7~cm from the requested goal, within the controller's 15~cm tolerance.

\subsection{Localization against a known map}
\begin{figure}[!t]
\centering
\includegraphics[width=\linewidth]{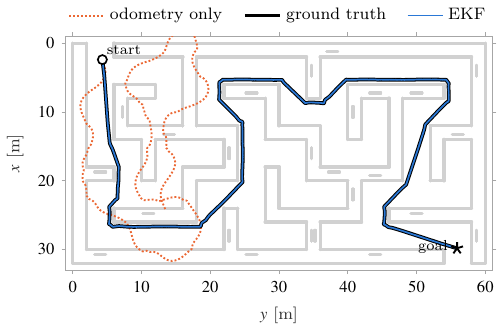}
\caption{Route from the start (left) to the goal (right), rotated to fit the column. The EKF estimate overlaps the ground truth at this scale; wheel odometry alone drifts through the walls and out of the maze.}
\label{fig:traj-ekf}
\end{figure}

\begin{figure}[!t]
\centering
\includegraphics[width=\linewidth]{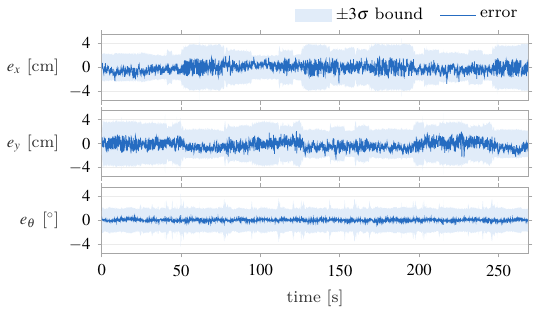}
\caption{EKF estimation error with $\pm 3\sigma$ bounds from the filter covariance.}
\label{fig:ekf-error}
\end{figure}

Fig.~\ref{fig:traj-ekf} shows the route and Fig.~\ref{fig:ekf-error} the estimation error. Over the whole run, the EKF position error has an RMSE of 1.0~cm and never exceeds 2.8~cm; the heading error has an RMSE of $0.16^\circ$ and a maximum of $0.83^\circ$. The errors stay within the filter's $\pm 3\sigma$ bounds 99.9\% of the time in $x$ and always in $y$ and $\theta$, so the filter is consistent and slightly conservative. Wheel odometry alone, integrated from the same start, ends 42.8~m from the true position. Most of that drift comes from a few sharp turns in which the wheels slide and odometry misses up to 40\% of the rotation; each such heading error bends the rest of the trajectory. The line updates remove it as it appears.

\subsection{EKF SLAM}
\begin{figure}[!t]
\centering
\includegraphics[width=\linewidth]{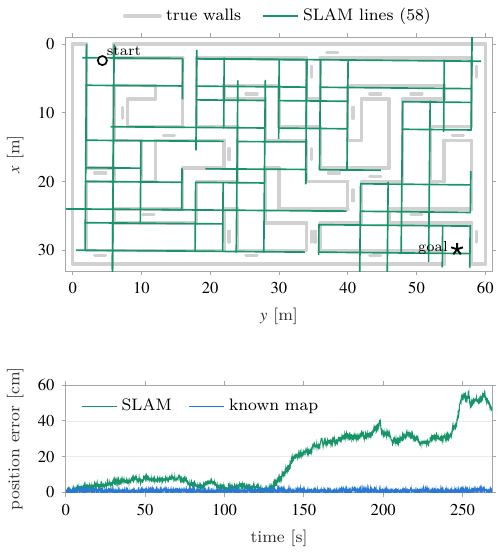}
\caption{EKF SLAM on the route of Fig.~\ref{fig:traj-ekf}. Top: final line map, each line drawn over its seen extent, over the true walls. Bottom: position error of SLAM and of localization against the known map.}
\label{fig:slam-map}
\end{figure}

In SLAM mode the robot drove the same route, from the same start, with no line map; the planner still used the known walls. Fig.~\ref{fig:slam-map} shows the map at the end of the run and the position error. The error stays below 10~cm for the first 136~s and then grows, to an RMSE of 24~cm over the run and a maximum of 56~cm; without loop closure, nothing corrects the error accumulated while new corridors are mapped. It nevertheless stays within the filter's $\pm 3\sigma$ bounds throughout, so the filter remains consistent. The robot reached the goal by its own estimate but stopped 0.52~m from it, the size of the drift. The NMPC was unaffected (10.8~ms mean solve time, no failed solves).

The map holds 58 lines for the 47 wall lines of the maze, so some walls were mapped more than once. Every mapped line matches a true wall line within $10^\circ$: the median offset error is 5.5~cm (95\% below 17.6~cm) and the median angle error $0.5^\circ$ (maximum $1.8^\circ$). The extents are less accurate than the lines. Only 56\% of the mapped extent lies within 0.3~m of a wall, and 74\% within 0.6~m: part of the difference is the drift, and part comes from the line extraction, whose RANSAC inliers are not split at gaps, so one extent can collect points from collinear walls in different corridors and bridge the openings between them, in one case beyond the maze boundary.

\subsection{Planning}
Table~\ref{tab:planners} compares the planners on the route of the experiments, on a $333 \times 612$ grid. A* finds the shortest grid path in 6.0~ms, but the path follows the grid directions and turns $2790^\circ$ in total. Shortcutting costs 0.1~ms more, removes 98\% of the points, and shortens the path to 133.8~m with a third of the turning, which is why it is the default. Hybrid A* yields a path of similar length that is smooth by construction, but takes 40 times longer, and only because of the obstacle-aware heuristic: with the Euclidean heuristic alone, it exhausts $2 \times 10^6$ expansions without reaching the goal. Measured every 5~cm, the closest approach to a wall is 0.65~m for A* and 0.60~m for the other paths, about the inflation radius.

\begin{table}[!t]
\centering
\caption{Planners from the start to the goal of the experiments (median of 20 runs; the last row is a single run)}
\label{tab:planners}
\setlength{\tabcolsep}{4pt}
\begin{tabular}{lrrrr}
\hline
Planner & Time [ms] & Expanded & Len.\ [m] & Turn [$^\circ$] \\
\hline
A* & 6.0 & 74\,500 & 136.5 & 2790 \\
A* + shortcut & 6.1 & 74\,500 & 133.8 & 958 \\
Hybrid A* & 242.7 & 239\,376 & 134.0 & 1452 \\
Hybrid A*, Eucl.\ $h$ & 2855 & $2 \times 10^6$ & \multicolumn{2}{l}{no path} \\
\hline
\end{tabular}
\end{table}

\subsection{Tracking and obstacle avoidance}
\begin{figure}[!t]
\centering
\includegraphics[width=\linewidth]{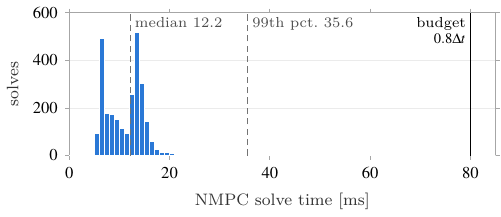}
\caption{NMPC solve times over a full run.}
\label{fig:solve-time}
\end{figure}

Along the route, the true position stays within 3.6~cm RMS of the planned path, with a maximum deviation of 17~cm. The NMPC ran 2688 solves (Fig.~\ref{fig:solve-time}) with a mean of 11.7~ms, a median of 12.2~ms, and a 99th percentile of 35.6~ms; none exceeded the CPU budget of $0.8\Delta t = 80$~ms, even measured as wall-clock time. A single solve did not converge, and the controller followed its previous plan for that step.

\begin{figure}[!t]
\centering
\includegraphics[width=\linewidth]{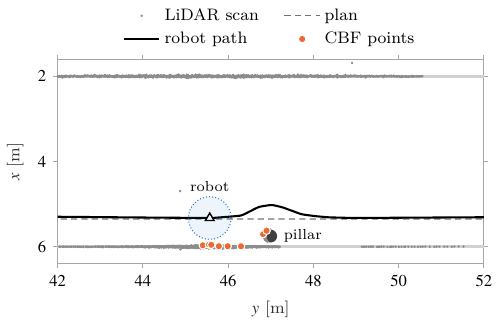}
\caption{Passing a pillar that is not in the map (dark disk), 0.4~m off the planned path. The triangle marks the robot 1.5~m before the pillar, with its 0.5~m safety radius (dotted); the orange points are the LiDAR points selected as CBF constraints at that instant, two of them on the pillar.}
\label{fig:obstacle}
\end{figure}

To test the CBF against an obstacle that is not in the map, we added a pillar of radius 0.15~m to the world as a separate model, so that neither the line map nor the planner's grid contains it, on the straight segment from $(5.35, 39.75)$ to $(5.35, 54.25)$~m of the route. With the pillar centered on the path, the robot stops 0.52~m from its surface and does not continue, although the corridor leaves 3.2~m free on one side. The tracking cost penalizes leaving the path to either side equally and the CBF forbids going through, so standing still is the cheapest solution: the constraints guarantee safety, not progress. With the pillar moved 0.4~m off the path toward the wall, so that the open side is the only way around it, the robot leaves the path by up to 0.33~m, passes the pillar with a minimum clearance of 0.57~m from its surface, and rejoins the path (Fig.~\ref{fig:obstacle}). Away from the pillar the deviation stays below 0.18~m, as in the run without it; the whole route takes 1~s longer, with no failed solves, and the EKF error is unaffected (1.1~cm RMSE).

\section{Conclusion and Future Work}

We presented Maze Runner, an open ROS~2 navigation stack for mazes and other environments made of planar walls. Localization relies only on the walls: lines extracted from the LiDAR scan correct wheel odometry in an EKF, against a map extracted from the maze mesh or, in SLAM mode, built online with candidate promotion and extent-aware association to cope with collinear walls. An earlier range--bearing model with point landmarks worked, but it needs beacons installed and maintained in the environment, which the line model avoids. An online grid planner provides a path in milliseconds, and an NMPC with discrete-time CBFs on world-frame LiDAR points tracks it while keeping clear of walls and unmapped obstacles. On a 135~m route in simulation, localization against the known map stays within 1.0~cm RMSE while wheel odometry alone drifts by more than 40~m; SLAM mode keeps the error at 24~cm RMSE without loop closure, within its $3\sigma$ bounds throughout; the planner returns a path in about 6~ms; and the NMPC solves in 11.7~ms on average, never over budget, while passing an unmapped pillar with 0.57~m clearance.

Several directions remain open. The planner only knows the walls in the map, so an unmapped obstacle centered on the path, or one that blocks the corridor, stops the robot safely but for good; feeding LiDAR obstacles back into the occupancy grid would let the planner route around it. The SLAM map is not yet used by the planner, which still requires a known maze. Complete coverage of the free space, which the earlier prototype addressed with a roadmap of Hybrid A* segments, is not yet part of the stack. Finally, the stack has only been evaluated in simulation; a bringup for the physical ROSbot XL, with recalibrated odometry, is the next step toward real-world deployment.

\bibliographystyle{IEEEtran}
\bibliography{refs}

\end{document}